\documentclass[11pt,a4paper]{article}
\usepackage[hyperref]{acl2021}
\usepackage{times}
\usepackage{latexsym}

\usepackage{microtype}

\aclfinalcopy % Uncomment this line for the final submission
\usepackage{graphicx} 
\usepackage{subfigure}
\usepackage{booktabs}
\usepackage{multirow}
\usepackage{amsmath} 
\usepackage{amssymb}           
\usepackage{amsfonts}            
\usepackage{mathrsfs}  
\usepackage{amsthm,amssymb}
\usepackage{color}
\usepackage{xcolor}%定义了一些颜色  
\usepackage{colortbl,booktabs}%第二个包定义了几个*rule 
\usepackage{fancybox}
\usepackage{framed} 
\usepackage{tikz}
\usepackage{pgfplots}
\usepackage{array}
\usepackage{microtype} 
\usepackage{tcolorbox} 
\usepackage[tikz]{bclogo}
\usepackage{lipsum} 
\usepackage{bbding}
\usepackage{pifont}
\usepackage{wasysym}
\usepackage{amssymb}
\usepackage[switch]{lineno}  %
\usepackage{pgfplots}
\pgfplotsset{width=7cm,compat=1.13}
\usepackage{subfigure}
\usepackage{url}
\usepackage{CJK}
\usepackage{ragged2e}
\title{Pre-training Universal Language Representation}

\author{Yian Li\textsuperscript{1,2,3}, Hai Zhao\textsuperscript{1,2,3,\thanks{\ \ Corresponding author. This paper was partially supported by National Key Research and Development Program of China (No. 2017YFB0304100), Key Projects of National Natural Science Foundation of China (U1836222 and 61733011), Huawei-SJTU long term AI project, Cutting-edge Machine Reading Comprehension and Language Model. This work was supported by Huawei Noah's Ark Lab.} } \\
\textsuperscript{1} Department of Computer Science and Engineering, Shanghai Jiao Tong University\\
\textsuperscript{2} Key Laboratory of Shanghai Education Commission for Intelligent Interaction\\
and Cognitive Engineering, Shanghai Jiao Tong University, Shanghai, China\\
\textsuperscript{3}MoE Key Lab of Artificial Intelligence, AI Institute, Shanghai Jiao Tong University\\
\texttt{liya19@sjtu.edu.cn,zhaohai@cs.sjtu.edu.cn}\\
}

\date{}

\begin{document}
\maketitle
\begin{abstract}
Despite the well-developed cut-edge representation learning for language, most language representation models usually focus on specific levels of linguistic units. This work introduces universal language representation learning, i.e., embeddings of different levels of linguistic units or text with quite diverse lengths in a uniform vector space. We propose the training objective MiSAD that utilizes meaningful $n$-grams extracted from large unlabeled corpus by a simple but effective algorithm for pre-trained language models. Then we empirically verify that well designed pre-training scheme may effectively yield universal language representation, which will bring great convenience when handling multiple layers of linguistic objects in a unified way. Especially, our model achieves the highest accuracy on analogy tasks in different language levels and significantly improves the performance on downstream tasks in the GLUE benchmark and a question answering dataset.
\end{abstract}

\section{Introduction}
In this paper, we propose universal language representation (ULR) that uniformly embeds linguistic units in different hierarchies in the same vector space. A universal language representation model encodes linguistic units such as words, phrases or sentences into fixed-sized vectors and handles multiple layers of linguistic objects in a unified way. ULR learning may offer a great convenience when confronted with sequences of different lengths, especially in tasks such as Natural Language Understanding (NLU) and Question Answering (QA), hence it is of great importance in both scientific research and industrial applications.
 
As is well known, embedding representation for a certain linguistic unit (i.e., word) enables linguistics-meaningful arithmetic calculation among different vectors, also known as word analogy \cite{word2vec}. For example:
\begin{equation*}
    \textit{King} - \textit{Man} = \textit{Queen} - \textit{Woman}
\end{equation*}
In fact, manipulating embeddings in the vector space reveals syntactic and semantic relations between the original symbol sequences and this feature is indeed useful in true applications. For example, ``\textit{London is the capital of England}” can be formulized as:
\begin{equation*}
    \textit{England} + \textit{capital} \approx \textit{London}
\end{equation*}
Then given two documents one of which contains ``\textit{England}” and ``\textit{capital}”, the other contains ``\textit{London}”, we consider them relevant. While a ULR model may generalize such good analogy features 
% or meaningful arithmetic operation 
onto free text with all language levels involved together. For example, \textit{Eat an onion} : \textit{Vegetable} :: \textit{Eat a pear} : \textit{Fruit}.

ULR has practical values in dialogue systems, by which human-computer communication will go far beyond executing instructions. One of the main challenges of dialogue systems is dialogue state tracking (DST). It can be formulated as a semantic parsing task \cite{dst}, namely, converting natural language utterances with any length into unified representations. Thus this is essentially a problem that can be conveniently solved by mapping sequences with similar semantic meanings into similar representations in the same vector space according to a ULR model. 

Another use of ULR is in the Frequently Asked Questions (FAQ) retrieval task, where the goal is to answer a user's question by retrieving question paraphrases that already have an answer from the database. 
% In such application, user's queries are paraphrases of the standard questions. 
Such task can be accurately done by only manipulating vectors such as calculating and ranking vector distance (i.e., cosine similarity). The core is to embed sequences of different lengths in the same vector space. 
% where similar sequences have similar latent representations. 
Then a ULR model retrieves the correct question-answer pair for the user query according to vector distance.

In this paper, we propose a universal language representation learning method that generates fixed-sized vectors for sequences of different lengths based on pre-trained language models \cite{bert, albert, electra}. We first introduce an efficient approach to extract and prune meaningful $n$-grams from unlabeled corpus. Then we present a new pre-training objective, Minimizing Symbol-vector Algorithmic Difference (MiSAD), that explicitly applies a penalty over different levels of linguistic units if their representations tend not to be in the same vector space.

To investigate our model's ability of capturing different levels of language information, we introduce an original universal analogy task derived from Google's word analogy dataset, where our model significantly improves the performance of previous pre-trained language models. Evaluation on a wide range of downstream tasks also demonstrates the effectiveness of our ULR model. Overall, our ULR-BERT obtains the highest average accuracy on our universal analogy dataset and reaches 1.1\% gain over Google BERT on the GLUE benchmark. Extensive experimental results on a question answering task verifies that our model can be easily applied to real-world applications in an extremely convenient way.

\section{Related Work}
% \subsection{Representation Models}
Previous language representation learning methods such as Word2Vec \cite{word2vec}, GloVe \cite{glove}, LASER \cite{laser}, InferSent \cite{infersent} and USE \cite{use} focus on specific granular linguistic units, e.g., words or sentences. Later proposed ELMo \cite{elmo}, OpenAI GPT \cite{gpt-1}, BERT \cite{bert} and XLNet \cite{xlnet} learns contextualized representation for each input token. Although such pre-trained language models (PrLMs) more or less are capable of offering universal language representation through their general-purpose training objectives, all the PrLMs devote into the contextualized representations from a generic text background and pay little attention on our concerned universal language presentation. 

As a typical PrLM, BERT is trained on a large amount of unlabeled data including two training targets: Masked Language Model (MLM), and Next Sentence Prediction (NSP). ALBERT \cite{albert} is trained with Sentence-Order Prediction (SOP) as a replacement of NSP. StructBERT \cite{structbert} combines NSP and SOP to learn inter-sentence structural information. Nevertheless, RoBERTa \cite{roberta} and SpanBERT \cite{spanbert} show that single-sequence training is better than the sentence-pair scenario. Besides, BERT-wwm \cite{bert-wwm}, StructBERT \cite{spanbert}, SpanBERT \cite{structbert} perform MLM on higher linguistic levels, augmenting the MLM objective by masking whole words, trigrams or spans, respectively. 
ELECTRA \cite{electra} further improves pre-training through a generator and discriminator architecture. The aforementioned models may seemingly handle different sized input sequences, but all of them focus on sentence-level specific representation still for each word, which may cause unsatisfactory performance in real-world situations.

There are a series of downstream NLP tasks especially on question-answering which may be conveniently and effectively solved through ULR like solution. Actually, though in different forms, these tasks more and more tend to be solved by our suggested ULR model, including dialogue utterance regularization \cite{yukai}, question paraphrasing \cite{qpr}, measuring QA similarities in FAQ tasks \cite{faq_transformer,faq_bert_1}.

\section{Model}
As pre-trained contextualized language models show their powerfulness in generic language representation for various downstream NLP tasks, we present a BERT-style ULR model that is especially designed to effectively learn universal, fixed-sized representations for input sequences of any granularity, i.e., words, phrases, and sentences. Our proposed pre-training method is furthermore strengthened in three-fold. First, we extract a large number of meaningful $n$-grams from monolingual corpus based on point-wise mutual information to leverage the multi-granular structural information. Second, inspired by word and phrase representation and their compositionality, we introduce a novel pre-training objective that directly models the extracted $n$-grams through manipulating their representations. Finally, we implement a normalized score for each $n$-gram to guide their sampling for training.

\subsection{$n$-gram Extracting}
Given a symbol sentence, \citet{spanbert} utilize span-level information by randomly masking and predicting contiguous segments. Different from such random sampling strategy, our method is based on point-wise mutual information (PMI) \cite{pmi} that makes efficient use of statistics and automatically extracts meaningful $n$-grams from unlabeled corpus.

Mutual information (MI) describes the association between two tokens by comparing the probability of observing them together with the probabilities of observing them independently. Higher mutual information indicates stronger association between the tokens. To be specific, an $\emph{n}$-gram is denoted as $w = (x_1, \ldots, x_{|w|})$, where $|w|$ is the number of tokens in $w$ and $n>1$. Therefore, we present an extended PMI formula displayed as below:
\begin{equation*}
 PMI(w) = \frac{1}{|w|} \left( {\rm log}P(w) - \sum\limits_{k=1}^{|w|}{\rm log}P(x_k) \right)
\end{equation*}
where the probabilities are estimated by counting the number of observations of each token and $n$-gram in the corpus, and normalizing by the corpus size. $\frac{1}{|w|}$ is an additional normalization factor which avoids extremely low scores for long $n$-grams.

We first collect all $\emph{n}$-grams with lengths up to $N$ using the SRILM toolkit\footnote{http://www.speech.sri.com/projects/srilm/download.html} \cite{srilm}, and compute PMI scores for all the $n$-grams based on their occurrences. Then, only $n$-grams with PMI scores higher than the chosen threshold are selected and input sequences are marked with the corresponding $n$-grams.

\subsection{Training Objective}
While the MLM training objective as in BERT \cite{bert} and its extensions \cite{bert-wwm, spanbert, structbert} are widely used for pre-trained contextualized language modeling, they do not focus on our concerned ULR, which demands an arithmetic corresponding relationship between the symbol and its represented vector. In order to directly model such demand, we propose a novel training target -- Minimizing Symbol-vector Algorithmic Difference (MiSAD) -- that leverages the vector space regularity of different granular linguistic units. For example, the following symbol sequence equation
\begin{align}
 &\textit{``London is"} + \textit{``the capital of England"} \notag\\
    = &\textit{``London is the capital of England"}     
  \label{eq1}
\end{align}
indicates a vector algorithmic equation according to our ULR goal, 
\begin{align}
 &vector(\textit{``London is"}) + vector(\textit{``the capital of} \notag\\
 &\textit{England"}) \notag\\
= &vector(\textit{``London is the capital of England"})       
  \label{eq2}
\end{align}
Thus, if the symbol equation (\ref{eq1}) cannot imply the respective vector equation (\ref{eq2}), we may set a training objective to let the ULR model forcedly learn such relationship.

Formally, we denote the input sequence by $S = \{x_1, \dots, x_m\}$, where $m$ is the number of tokens in $S$. After $n$-gram extracting and pruning by means of PMI, each sequence is marked with several $n$-grams. During pre-training, only one of them is selected by the $n$-gram scoring function, which will be introduced in detail in Section \ref{score}, and the input sequence is represented as $S = \{x_1, \dots, x_{i-1}, w, x_{j+1}, \dots, x_m\}$, where the $n$-gram $w = \{x_i, \dots, x_j\}$ ($1 \le i < j \le m$) is a sub-sequence of $S$. Then we convert $S$ into two independent parts -- the $n$-gram $w$ and the rest of the tokens $R = \{x_1, \dots, x_{i-1}, x_{j+1}, \dots, x_m\}$ -- which are fed into the model separately along with the original complete sequence.

The Transformer encoder generates a contextualized representation for each token in the sequence. To derive fixed-sized vectors for sequences of different lengths, we use the pooled output of the \texttt{[CLS]} token as sequence embeddings. The model is trained to minimize the following Mean Square Error (MSE) loss:
\begin{equation*}
 \mathcal{L}_{MiSAD} = MSE(E^w + E^R, E^S)
\end{equation*}
where $E^w$, $E^R$ and $E^S$ are representations of $w$, $R$ and $S$, respectively, and are all normalized to unit lengths. 
To enhance the robustness of the model, we jointly train MiSAD and the MLM objective $\mathcal{L}_{MLM}$ as in BERT with equal weights. Since the input sentence $S$ is split into $w + R$, we must avoid masking out the $n$-gram $w$ in the original sentence in order not to affect the semantics after vector space combination. However, tokens in $n$-grams other than $w$ have equal weights of being replaced with \texttt{[MASK]} as other tokens. The final loss function is as follows:
\begin{equation*}
 \mathcal{L} = \mathcal{L}_{MiSAD} + \mathcal{L}_{MLM}
\end{equation*}

\subsection{$n$-gram Sampling}
\label{score}
For a given sequence, the importance of different $n$-grams and the degree to which the model understands their semantics are different. Instead of sampling $n$-grams at random, we let the model decide which $n$-gram to choose based on the knowledge learned in the pre-training stage. Following \citet{ngram_scoring}, we employ a normalized score for each $n$-gram in the input sequence using the masked language modeling head.

We mask one $n$-gram at a time and the model outputs probabilities of the masked tokens given their surrounding context. The score of an $n$-gram $w$ is calculated as the average probabilities of all tokens in it.
\begin{equation*}
 score_w = \frac{1}{|w|} \sum^{|w|}_{k=1} P(x_k|S^{\backslash w})
\end{equation*}
where $|w|$ is the length of $w$ and $S^{\backslash w}$ is the notation of an input sequence $S$ with each token within the $n$-gram $w$ replaced by the special token \texttt{[MASK]}. Finally, we choose the \textit {n}-gram with the lowest score for our training target.

\section{Implementation of ULR Pre-training}
This section introduces our ULR pre-training details.

As for the pre-training corpus, we download the English Wikipedia Corpus\footnote{https://dumps.wikimedia.org/enwiki/latest} and pre-process with \texttt{process\_wiki.py}\footnote{https://github.com/panyang/Wikipedia\_Word2vec/blob/\\master/v1/process\_wiki.py}, which extracts text from xml files. When processing paragraphs from Wikipedia, we find that a large number of entities are annotated with special marks, which may be useful for our task. Therefore, we identify all the entities and treat them as high-quality $n$-grams. Then, we remove punctuation marks and characters in other languages based on regular expressions, and finally get a corpus of 2,266M words. 

As for $n$-gram pruning, PMI scores of all $n$-grams with a maximum length of $N=6$ are calculated for each document. We manually evaluate the extracted $n$-grams and find more than 50\% of the top 2000 $n$-grams contain 2 $\sim$ 3 words, and only less than 3\% $n$-grams are longer than 4. Although a larger $n$-gram vocabulary can cover longer $n$-grams, it will cause too many meaningless $n$-grams at the same time. Therefore, we empirically retain the top 3000 $n$-grams for each document. Finally, we randomly sample 10M sentences from the entire corpus to reduce training time.

During pre-training, BERT packs sentence pairs into a single sequence and use the special \texttt{[CLS]} token as sentence-pair representation. However, our MiSAD training objective requires single-sentence inputs. Thus in our experiments, each input is an $n$-ngram or a single sequence with a maximum length of 128. Special tokens \texttt{[CLS]} and \texttt{[SEP]} are added at the front and end of each input, respectively. Instead of training from scratch, we initialize our model with the officially released checkpoints of BERT \cite{bert}, ALBERT \cite{albert} and ELECTRA \cite{electra}. 
% The base model consists of 12 Transformer layers, 12 heads, 768 dimensional hidden states and 110M parameters in total and the large model has 24 Transformer layers, 16 heads, 1024 dimensional hidden states and 340M parameters in total. 
We use Adam optimizer \cite{adam} with initial learning rate of 5e-5 and linear warmup over the first 10\% of the training steps. Batch size is 64 and dropout rate is 0.1. Each model is trained for one epoch over 10M training examples on four Nvidia Tesla P40 GPUs. 

To derive fixed-dimensional vectors of input sequences, we apply three pooling strategies on top of the PrLM: Using the vector of the \texttt{[CLS]} token, mean-pooling of all token embeddings and max-pooling over time of all embeddings. The default setting is mean-pooling. 

\begin{table*}[t]
  \centering
  \small
%   \resizebox{\textwidth}{!}{
  \begin{tabular}{lllllllllll}\toprule
    %  A &$+$& B &$-$& C &$=$& \multicolumn{5}{c}{D}\\\midrule
    \textit{girl}&$-$& \textit{boy}&$+$&\textit{brother} &$=$& \textit{daughter} & \textit{\textbf{sister}} & \textit{wife} & \textit{father} & \textit{son}\\\hline
    
     \textit{worse}&$-$&\textit{bad}&$+$&\textit{big} &$=$& \textit{\textbf{bigger}} &\textit{larger} & \textit{smaller} & \textit{biggest} &\textit{better}\\\hline
     
     \textit{China}&$-$&\textit{Beijing}&$+$&\textit{Paris} &$=$& \textit{\textbf{France}} & \textit{Europe} & \textit{Germany} & \textit{Belgium} & \textit{London}\\
     \hline
     
     \textit{Chilean}&$-$&\textit{Chile}&$+$&\textit{China} &$=$& \textit{Japanese} & \textit{\textbf{Chinese}} & \textit{Russian} & \textit{Korean} & \textit{Ukrainian}\\\bottomrule
  \end{tabular}
%   }
\caption{Examples from our word analogy dataset. The correct answers are in bold.}
  \label{word-analogy}
\end{table*}

\section{Experimental Setup}
\subsection{Tasks}
We construct a universal analogy dataset in terms of words, phrases and sentences and experiment with multiple representation models to examine their ability of representing different levels of linguistic units through a task-independent evaluation\footnote{Code and dataset are available at: https://github.com/\\Liyianan/ULR.}. Furthermore, we conduct experiments on a wide range of downstream tasks from the GLUE benchmark and a question answering task. 

\subsubsection{Universal Analogy}
Our universal analogy dataset is based on Google's word analogy dataset and contains three levels of tasks: words, phrases and sentences. 

\noindent
\textbf{Word-level} Recall that in a word analogy task \cite{word2vec}, two pairs of words that share the same type of relationship, denoted as $A$ : $B$ :: $C$ : $D$, are involved. The goal is to retrieve the last word from the vocabulary given the first three words. To facilitate comparison between models with different vocabularies, we construct a closed-vocabulary analogy task based on Google's word analogy dataset through negative sampling. Concretely, for each original question, we use GloVe to rank every word in the vocabulary and the top 5 results are considered to be candidate words. If GloVe fails to retrieve the correct answer, we manually add it to make sure it is included in the candidates. During evaluation, the model is expected to select the correct answer from 5 candidate words. Table \ref{word-analogy} shows examples from our word anlogy dataset. 

\noindent
\textbf{Phrase-/Sentence-level}
To derive higher level analogy datasets, we put word pairs from the word-level dataset into contexts so that the resulting phrase and sentence pairs also have linear relationships. Phrase and sentence templates are extrated from the English Wikipedia Corpus. Both phrase and sentence datasets have four types of semantic analogy and three kinds of syntactic analogy. Please refer to Appendix A for details about our approach of constructing the universal analogy dataset.

\begin{table*}[t]
\small
  \centering
  \renewcommand{\multirowsetup}{\centering}
  \begin{tabular}{l|ccc|ccc|ccc|cc}\toprule
  \multirow{2}{*}{\textbf{Model}}
  & \multicolumn{3}{c|}{\textbf{Word}} & \multicolumn{3}{c|}{\textbf{Phrase}} &
  \multicolumn{3}{c|}{\textbf{Sentence}}&\multirow{2}{*}{\textbf{Avg.}}& \multirow{2}{*}{\textbf{Gain}} \\
  \cline{2-10} \rule{0pt}{12pt}
    & sem & syn & Avg. &sem & syn & Avg. & sem & syn & Avg. & \\\midrule
    \multicolumn{4}{l}{\textit{Word \& Sentence Representation Models}}\\
    GloVe  & \textbf{82.6} & 78.0	& \textbf{80.3}	& \;\;0.0	& 40.9	& 20.5	& \;\;0.2 &	39.8 &	20.0 &	40.3  &-\\
    
    InferSent & 68.8 &	\textbf{88.7} &	78.8&	\;\;0.0 &	54.1 &	27.0 &	\;\;0.0 &	50.8 &	25.4 &	43.7&-\\
    GenSen& 44.5 &	84.4 &	64.5 &	\;\;0.0 &	54.4 &	27.2 &	\;\;0.0 &	44.9 &	22.4 &	38.0&- \\
    USE & 73.0 &	83.1 &	78.0 &	\;\;1.8 &	63.1 &	32.5 &	\;\;0.6 &	44.1 &	22.4 &	44.3 &-\\
    LASER  &26.9 & 78.2&52.6 &\;\;0.0 &63.3 &31.7 & \;\;1.6 & 55.4 & 28.5 &37.6 &-\\ \midrule
    
    \multicolumn{4}{l}{\textit{Pre-trained Contextualized Language Models}}\\
    BERT\textsubscript{BASE}& 51.3 &	60.2 &	55.8 &\;\;0.3 & 	\textbf{69.3} &\textbf{34.8}	 &\;\;0.1 &	\textbf{68.3} &\textbf{34.2}	&41.6	&- \\
    BERT\textsubscript{LARGE} & 49.7 &	46.6 &	48.2 &\;\;0.1 & 67.4&33.9 &\;\;0.5	 &61.2 	&30.9& 37.7&-\\
    ALBERT\textsubscript{BASE} & 33.7	& 38.1	& 35.9	& \;\;0.1	& 53.6	& 26.7 &	\;\;0.1	& 60.9	& 30.5	& 31.0 &-\\
    ALBERT\textsubscript{XXLARGE} & 38.2	&35.6	&36.9	&\;\;0.8	&	52.3&26.6	&\;\;0.4	&49.4	&24.9	&29.5&-\\
    ELECTRA\textsubscript{BASE} &22.9&32.4&27.7&\;\;2.2&57.1&29.7&\;\;0.4&39.5&20.0&25.8 &-\\
    ELECTRA\textsubscript{LARGE} &20.4&24.7&22.6& \;\;2.9&49.8&26.4&\;\;1.4&52.0&26.7&25.2&-\\\midrule
    
    \multicolumn{4}{l}{\textit{Our Universal Language Representation Models}}\\
    ULR-BERT\textsubscript{BASE}& 71.7& 70.0& 70.8 & \;\;1.1& 66.8&34.0 & \;\;1.5& 63.0& 32.3& 45.7&4.1 \\
    ULR-BERT\textsubscript{LARGE}&80.8& 66.2& 73.5& \;\;\textbf{8.4}& 60.5& 34.5& \;\;\textbf{4.7}& 54.3& 29.5&\textbf{45.8}&\textbf{8.1} \\
    ULR-ALBERT\textsubscript{BASE}&43.5& 56.3& 49.9& \;\;0.3& 58.2& 29.3& \;\;0.3& 60.9& 30.6&36.6&5.6 \\
    ULR-ALBERT\textsubscript{XXLARGE}&26.8& 31.0& 28.9& \;\;3.6& 55.0& 29.3& \;\;0.7& 60.3& 30.5& 29.6&0.1\\
    ULR-ELECTRA\textsubscript{BASE}&24.4& 34.6&29.5& \;\;1.7& 56.5& 29.1& \;\;0.9& 57.6& 29.3&29.3&3.5\\
    ULR-ELECTRA\textsubscript{LARGE}&22.0& 31.0&26.5& \;\;2.9& 56.7&29.8& \;\;0.8& 52.9&26.9&27.7&2.5\\\bottomrule
  \end{tabular}
     \caption{Performance of different models on the universal analogy dataset. ``sem" = semantic. ``syn" = syntactic.}
  \label{analogy}
\end{table*}

\subsubsection{GLUE}
The General Language Understanding Evaluation (GLUE) benchmark \cite{glue} is a collection of tasks that are widely used to evaluate the performance of a model in language understanding. We divide NLU tasks from the GLUE benchmark into three main categories.

\noindent
\textbf{Single-Sentence Classification} Single-sentence classification tasks includes SST-2 \cite{sst-2}, a sentiment classification task, and CoLA \cite{cola}, a task that is to determine whether a sentence is grammatically acceptable. 

\noindent
\textbf{Natural Language Inference} GLUE contains four NLI tasks: MNLI \cite{mnli}, QNIL \cite{qnli}, RTE \cite{rte} and WNLI \cite{wnli}. However, we exclude the problematic WNLI in accordance with \citet{bert}.

\noindent
\textbf{Semantic Similarity} MRPC \cite{mrpc}, QQP \cite{qqp} and STS-B \cite{sts-b} are semantic similarity tasks, where the model is required to either determine whether the two sentences are equivalent or assign a similarity score for them.

In the fine-tuning stage, pairs of sentences are concatenated into a single sequence with a special token \texttt{[SEP]} in between. For both single sentence and sentence pair tasks, the hidden state of the first token \texttt{[CLS]} is used for softmax classification. We use the same sets of hyperparameters for all the evaluated models. Experiments are ran with batch sizes in \{8, 16, 32, 64\} and learning rate of 3e-5 for 3 epochs. 

\subsubsection{\textsc{GeoGranno}}
\textsc{GeoGranno} \cite{geogranno} contains natural language paraphrases paired with 
% their 
logical forms. The dataset is manually annotated: For each natural language utterance, a correct canonical utterance paraphrase is selected. The train/dev sets have 487 and 59 paraphrase pairs, respectively. In our experiments, we focus on question paraphrase retrieval, whose task is to retrieve the correct paraphrase from all 158 different sentences when given a question. Most of the queries have only one correct answer while some have two or more matches. Evaluation metrics are Top-1/5/10 accuracy.

\subsection{Baselines}
On the universal analogy task, we adopt three types of baselines including bag-of-words (BoW) model from pre-trained word embeddings: GloVe \cite{glove}, sentence embedding models: InferSent \cite{infersent}, GenSen \cite{gensen}, USE \cite{use} and LASER \cite{laser}, and pre-trained contextualized language models: BERT, ALBERT and ELECTRA. 

On GLUE and \textsc{GeoGranno}, we especially evaluate our model and two baseline models: 

\paragraph{BERT} The officially released pre-trained BERT models \cite{bert}.

\paragraph{MLM-BERT} BERT models trained with the same additional steps with our model on Wikipedia using only the MLM objective.

\paragraph{ULR-BERT} Our universal language representation model trained on Wikipedia with MLM and MiSAD.

\begin{table*}[t]{}
  \centering
  \small
%   \resizebox{0.98\textwidth}{!}{
  \begin{tabular}{lcccccccccccc}
  
  \multicolumn{12}{c}{\textit{Batch size: 8, 16, 32, 64; Length: 128; Epoch: 3; lr: 3e-5}}\\ \toprule
  
  \multirow{3}{*}{\textbf{Model}}
  & \multicolumn{2}{c}{\textbf{Single Sentence}} && \multicolumn{3}{c}{\textbf{Natural Language Inference}} &&
  \multicolumn{3}{c}{\textbf{Semantic Similarity}}&\multirow{3}{*}{\textbf{Avg.}} &\multirow{3}{*}{\textbf{Gain}} \\
  \cline{2-3} \cline{5-7} \cline{9-11} \rule{0pt}{12pt}
    & CoLA & SST-2&& MNLI & QNLI & RTE && MRPC & QQP & STS-B & \\
    & (mc) & (acc) && m/mm(acc) & (acc) & (acc) && (F1) & (F1) & (pc)  & \\\midrule
    \multicolumn{12}{c}{\textit{In literature}}\\ 
    BERT\textsubscript{BASE} 	&52.1 &93.5 && 84.6/83.4 &90.5&	66.4 && 88.9	&	71.2& 87.1& 79.7&-\\
    BERT\textsubscript{LARGE}  & 60.5	&94.9 &&\textbf{86.7}/85.9 &92.7&	70.1&& 89.3	&	72.1& 87.6& 82.2&-\\\midrule
    \multicolumn{12}{c}{\textit{Our implementation}}\\
    BERT\textsubscript{BASE} & 53.5 & 92.1 && 84.5/83.7 & 90.6 & 67.1 && 87.5 & 71.6 & 85.3 & 79.5&-\\
    MLM-BERT\textsubscript{BASE}& 51.9 & 94.0 && 84.5/83.9 & 90.4 & 66.6 && 88.1 & 71.6 & 86.2 &79.7&0.2\\
    ULR-BERT\textsubscript{BASE} & 56.5 & 94.3 && 84.6/84.0 & 91.0 & 68.0 && 89.0 & 71.6 & 86.6 &80.6&1.1\\
    BERT\textsubscript{LARGE}  & 60.5 & 94.9 && 86.1/85.6 & 92.8 & 68.8 && 89.6 & 72.1 & 87.3 &82.0&-\\
    MLM-BERT\textsubscript{LARGE}& \textbf{62.6} & 94.5 && 86.6/85.6 & 92.8 & 67.1 && 88.9 & \textbf{72.3} & 87.2 &82.0&\;\;\;0\\
    ULR-BERT\textsubscript{LARGE} & 61.8 & \textbf{95.0} && \textbf{86.7}/\textbf{86.0} & \textbf{93.0} & \textbf{71.0} && \textbf{90.2} & \textbf{72.3} & \textbf{88.2} &\textbf{82.7}&0.7\\\bottomrule
  \end{tabular}
%   }
\caption{Test results on the GLUE benchmark scored by the evaluation server\protect\footnotemark. We exclude the problematic WNLI dataset and recalculate the ``Avg." score. Results for BERT\textsubscript{BASE} and BERT\textsubscript{LARGE} are obtained from \citet{bert}. ``mc" and ``pc" are Matthews correlation coefficient \cite{mc} and Pearson correlation coefficient, respectively.}
  \label{glue}
\end{table*}
\footnotetext{https://gluebenchmark.com}

\begin{table}[t]{}
  \centering
  \small
%   \resizebox{0.98\columnwidth}{!}{
  \begin{tabular}{lccc} \toprule
  \textbf{Model} & \textbf{Top-1} & \textbf{Top-5} & \textbf{Top-10} \\\midrule 
  GloVe & \;\;0.3 & \;\;2.7 & \;\;7.4\\
  LASER & \;\;6.3 & \;\;9.5 & 12.7 \\
%   TF-IDF &24.4 & 61.6& 78.1 \\
  BM25 & 27.1 & 62.5 & 76.4 \\\midrule
  
  BERT\textsubscript{BASE} & 29.6 & 58.9 & 67.1\\
  MLM-BERT\textsubscript{BASE} &37.0 & 66.8 & 72.6\\
  ULR-BERT\textsubscript{BASE} & \textbf{39.7 }& 66.0 & \textbf{77.3}\\
  BERT\textsubscript{LARGE} & 15.9 &42.7 & 54.2\\
  MLM-BERT\textsubscript{LARGE}& 24.5 & 57.8 & 70.7 \\
  ULR-BERT\textsubscript{LARGE} & 35.1 & \textbf{68.8 }& \textbf{77.3}\\\bottomrule
  \end{tabular}
%   }
  \caption{Question paraphrase retrieval accuracy of different models on the train-dev set of \textsc{GeoGranno}.}
  \label{geo}
\end{table}

\section{Results}
\subsection{Universal Analogy}
Results on our universal analogy dataset are reported in Table \ref{analogy}. Generally, semantic analogies are more challenging than the syntactic ones and higher-level relationships between sequences are more difficult to capture, which is observed in almost all the evaluated models. On the word analogy task, GloVe achieves the highest accuracy (80.3\%) while its performance drops sharply on higher-level tasks. All well trained PrLMs like BERT, ALBERT and ELECTRA hardly exhibit arithmetic characteristics and increasing the model size usually leads to a decrease in accuracy.

However, training models with our properly designed MiSAD objective greatly improves the performance, especially in word-level analogy. Especially, ULR-BERT obtains 15\% $\sim$ 25\% absolute gains, such results are so strong to be comparable to GloVe, which especially focuses on the linear word analogy feature from its training scheme. Meanwhile GloVe performs far worse than our model on higher-level analogies. Overall, ULR-BERT achieves the highest average accuracy (45.8\%), an absolute gain of 8.1\% over BERT, indicating that it has indeed more effectively learned universal language representations across different linguistic units. It demonstrates that our pre-training method is effective and can be adapted to different PrLMs.

\subsection{GLUE}
Table \ref{glue} shows the performance on the GLUE benchmark. Our model improves the BERT\textsubscript{BASE} and BERT\textsubscript{LARGE} by 1.1\% and 0.7\% on average, respectively. Since our model is established on the released checkpoints of Google BERT, we make additional comparison with MLM-BERT that is trained under the same procedure as our model except for the pre-training objective. While the model trained with more MLM updates may improve the performance on some tasks, it underperforms BERT on datasets such as MRPC, RTE and SST-2. Our model exceeds MLM-BERT\textsubscript{BASE} and MLM-BERT\textsubscript{LARGE} by 0.9\% and 0.7\% on average respectively. The main gains from the base model are in CoLA (+4.6\%) and RTE (+1.4\%), which are entirely contributed by our MiSAD training objective. Overall, our model improves the performance of its baseline on every dataset in the GLUE benchmark, demonstrating its effectiveness in real applications of natural language understanding. 

\subsection{\textsc{GeoGranno}}
Table \ref{geo} shows the performance on \textsc{GeoGranno}. As we can see, 4 out of 6 evaluated pre-trained language models significantly outperform BM25 for Top-1 accuracy, indicating the superiority of embedding-based models over the statistical method. Among all the evaluated models, our ULR-BERT yields the highest accuracies (39.7\%/68.8\%/77.3\%). To be specific, our ULR model exceeds BERT by 10.1\% and 19.2\% Top-1 accuracy and obtains 2.7\% and 10.6\% improvements compared with MLM-BERT, which are consistent with the results on the GLUE benchmark. Since $n$-grams and sentences of different lengths are involved in the pre-training of our model, it is especially better at understanding the semantics of input sequences and mapping queries to their paraphrases according to the learned sense of semantic equality.

\section{Ablation Study}
In this section, we explore to what extent does our model benefit from the MiSAD objective and sampling strategy, and further confirm that our pre-training procedure improves the model's ability of encoding variable-length sequences. 

\begin{table*}[t]{}
  \centering
  \small
%   \resizebox{0.98\textwidth}{!}{
  \begin{tabular}{lcccccccccccc}  \toprule
  
  \multirow{3}{*}{\textbf{Model}}
  & \multicolumn{2}{c}{\textbf{Single Sentence}} && \multicolumn{3}{c}{\textbf{Natural Language Inference}} &&
  \multicolumn{3}{c}{\textbf{Semantic Similarity}}&\multirow{3}{*}{\textbf{Avg.}}&\multirow{3}{*}{\textbf{Gain}}\\
  \cline{2-3} \cline{5-7} \cline{9-11} \rule{0pt}{12pt}
    & CoLA & SST-2&& MNLI & QNLI & RTE && MRPC & QQP & STS-B & \\
    & (mc) & (acc)&& m/mm(acc) & (acc) & (acc) && (F1) & (F1) & (pc)  &
     \\\midrule
    BERT & 53.5 & 92.1 && 84.5/83.7 & 90.6 & 67.1 && 87.5 & 71.6 & 85.3 & 79.5&-\\
    MLM-BERT&  51.9 & 94.0 && 84.5/83.9 & 90.4 & 66.6 && 88.1 & 71.6 & 86.2 &79.7&\;0.2\\

    NSP-BERT & 53.5 & 93.2 && 84.1/83.5 & 90.5 & 66.1 && 87.7 & \textbf{72.1} & 84.5 & 79.5&\;\;\;\;0\\ 
    SOP-BERT & 50.9 & 92.7 && 84.0/83.1 & 90.7 & 66.5&& 85.0 & 70.9& 83.9 & 78.6&-0.9\\
    
    ULR-BERT & \textbf{56.5} & \textbf{94.3} && \textbf{84.6}/\textbf{84.0} & \textbf{91.0} & \textbf{68.0} && \textbf{89.0} & 71.6 & \textbf{86.6} &\textbf{80.6}&\;1.1\\\bottomrule
  \end{tabular}
%   }
  \caption{Comparison of base models using different training objectives on the GLUE test set.}
  \label{ablation-1}
\end{table*}

\begin{table*}[t]{}
  \centering
  \small
%   \resizebox{0.98\textwidth}{!}{
  \begin{tabular}{lccccccccccc}  \toprule
  
  \multirow{2}{*}{\textbf{Model}}
  & \multicolumn{3}{c}{\textbf{CoLA}} && \multicolumn{3}{c}{\textbf{RTE}} &&
  \multicolumn{3}{c}{\textbf{MRPC}}\\
%   \cline{2-3} \cline{5-7} \cline{9-11} \rule{0pt}{12pt}
    % & CoLA & SST-2&& MNLI & QNLI & RTE && MRPC & QQP & STS-B & \\
    & std. & mean & max&& std. & mean & max&&std. & mean & max
     \\\midrule
    BERT & 1.51  &57.0 & 58.3&&	1.92&68.1  &70.4&&	0.52 & 90.4 & 90.9\\
    ULR-BERT&  1.31 & 59.3 & 60.2&&	1.83	& 69.3&  72.6&&	0.61&  90.8&  91.5\\
    \bottomrule
  \end{tabular}
%   }
  \caption{Standard deviation, mean, and maximum performance on the GLUE dev set when fintuing BERT and ULR-BERT with 5 random seeds.}
  \label{rebuttal-1}
\end{table*}

\subsection{Effect of Training Objectives}
To make a fair comparison, we train BERT with the same additional updates using different combinations of training tasks:

\paragraph{NSP-BERT} is trained with MLM and NSP, whose goal is to distinguish whether two input sentences are consecutive. For each sentence, we choose its following sentence 50\% of the time and randomly sample a sentence 50\% of the time. 

\paragraph{SOP-BERT} is trained with MLM and SOP, a substitute of the NSP task that aims at better modeling the coherence between sentences. Consistent with \citet{albert}, we sample two consecutive sentences in the same document as a positive sample, and reverse their order 50\% of the time to create a negative sample.

\paragraph{}For both baselines and ULR, we use the same set of parameters for 5 runs, and average scores on the GLUE test set are reported in Table \ref{ablation-1}. Although we expect NSP and SOP to help the model better understand the relationship between sentences and benefit tasks like natural language inference, they hardly improve the performance on GLUE according to our strict implementation. Specifically, NSP-BERT outperforms MLM-BERT on datasets such as CoLA, QNLI and QQP while less satisfactory on other tasks. SOP-BERT is on a par with MLM-BERT on three NLI tasks but it sharply decreases the score on other datasets. In general, single-sentence training with only the MLM objective accounts for better performance as described by \citet{albert, roberta}. Besides, our training strategy which combines MLM and MiSAD yields the most considerable gains compared with other training objectives.

Table \ref{rebuttal-1} shows standard deviation, mean and maximum performance on CoLA/RTE/MRPC dev set when fine-tuning BERT and ULR-BERT over 5 random seeds, which clearly shows that our model is generally more stable and yields better results compared with BERT.

\subsection{Effect of Sampling Strategies}
We compare our PMI-based $n$-gram sampling scheme with two alternatives. Specifically, we train the following two baseline models under the same model settings except for the sampling strategy.

\noindent
\textbf{Random Spans} We replace our $n$-gram module with the masking strategy as proposed by \citet{spanbert}, where the sampling probability of span length $l$ is based on a geometric distribution $l \sim Geo(p)$. The parameter $p$ is set to 0.2 and maximum span length $l_{max}=6$. 

\noindent
\textbf{Named Entities} 
We only retain named entities that are annotated in the Wikipedia Corpus. 
% The average length of all named entities in our dataset is {}.

Table \ref{ablation-2} shows the effect of different sampling schemes on the GLUE dev set. As we can see, our PMI-based $n$-gram sampling is preferable to other strategies on 6 out of 8 tasks. CoLA and RTE are more sensible to sampling strategies than other tasks. On average, using named entities and meaningful $n$-grams is better than randomly sampled spans. We attribute the source to the reason is that random span sampling ignores important semantic and syntactic structure of a sequence, resulting in a large number of meaningless segments. Compared with using only named entities, our PMI-based approach automatically discovers structures within any sequence and is not limited to any granularity, which is critical to pre-training universal language representation.  

\begin{table*}[t]{}
  \centering
  \small
%   \resizebox{0.98\textwidth}{!}{
  \begin{tabular}{lccccccccccc} \toprule
  
  \multirow{3}{*}{\textbf{Model}}
  & \multicolumn{2}{c}{\textbf{Single Sentence}} && \multicolumn{3}{c}{\textbf{Natural Language Inference}} &&
  \multicolumn{3}{c}{\textbf{Semantic Similarity}}&\multirow{3}{*}{\textbf{Avg.}}\\
  \cline{2-3} \cline{5-7} \cline{9-11} \rule{0pt}{12pt}
    & CoLA & SST-2&& MNLI & QNLI & RTE && MRPC & QQP & STS-B & \\
    & (mc) & (acc)&& m/mm(acc) & (acc) & (acc) && (F1) & (F1) & (pc)  &
     \\\midrule
    Random Spans &56.1 & 93.1 && 84.5/\textbf{84.9} & 91.5 & 66.1 && \textbf{91.5} & \textbf{87.9} & 89.8 & 82.8\\
    Named Entities & 57.1 & 93.1 && 84.4/84.7 & 91.6 & 67.5 && 90.8 & \textbf{87.9}& \textbf{89.9} & 83.0\\ 
    PMI $n$-grams & \textbf{59.3} & \textbf{93.6} && \textbf{84.7}/\textbf{84.9} & \textbf{91.8} & \textbf{69.3} && 90.8 & 87.8 & \textbf{89.9} & \textbf{83.6}\\
    \bottomrule
  \end{tabular}
%   }
  \caption{Comparison of base models using different sampling strategies on the GLUE dev set.}
  \label{ablation-2}
\end{table*}

\begin{table}[t]{}
  \centering
  \small
%   \resizebox{0.98\columnwidth}{!}{
  \begin{tabular}{lcc} \toprule
  \textbf{Model} & \textbf{\textsc{GeoGranno}} & \textbf{GLUE dev} \\\midrule 
  BERT  &29.6/58.9/67.1 & 82.6\\
  ALBERT & 18.4/41.1/52.6 & 83.0\\
  ELECTRA &11.2/21.1/26.6 & 86.5\\\midrule
  ULR-BERT &39.7/66.0/77.3 &83.6\\
  ULR-ALBERT &24.9/44.7/55.9 & 83.4 \\
  ULR-ELECTRA &26.8/51.8/65.5 &86.9\\\bottomrule
  \end{tabular}
%   }
  \caption{Comparison of different base models on \textsc{GeoGranno} and GLUE. We report the Top-1/5/10 accuracy on \textsc{GeoGranno}.}
  \label{ablation-3}
\end{table}

\subsection{Application to Different Models}
Experiments on the universal analogy task reveal that our proposed training scheme can be adapted to various pre-trained langauge models. In this subsection, we compare our model with BERT, ALBERT and ELECTRA on \textsc{GeoGranno} and the GLUE benchmark.

Table \ref{ablation-3} shows the results on \textsc{GeoGranno} and the GLUE dev set, where our approach can enhance the performance of all three pre-trained models. Among all the evaluated models, ULR-BERT achieves the largest gains on GLUE while ULR-ELECTRA obtains the most significant improvement on \textsc{GeoGranno}. It further verifies the effectiveness and universality of our model.

\begin{table}[t]{}
  \centering
  \small
%   \resizebox{0.98\columnwidth}{!}{
  \begin{tabular}{lccc} \toprule
  \multicolumn{4}{l}{\textit{Group by query length $|q|$}}\\ 
  \multirow{2}{*}{\textbf{Model}} & 1$\sim$6 & 7$\sim$8 &9$\sim$15 \\
  &(32.6\%)&(36.7\%)&(30.7\%)\\\midrule 
  BERT  &73.9&\;\;64.9&\;\;62.5\\
  ULR-BERT &79.8&\;\;76.1&\;\;75.9\\
  & +5.9& +11.2& +13.4\\\midrule
  \multicolumn{4}{l}{\textit{Group by $abs(|q|-|Q|)$}}\\ 
  \multirow{2}{*}{\textbf{Model}} & $\geq$0 & $\geq$2 &$\geq$3 \\
  &(100\%)&(62.2\%)&(43.3\%)\\\midrule
  BERT  &\;\;67.1 & \;\;63.0& \;\;57.0\\
  ULR-BERT &\;\;77.3 &\;\;76.2& \;\;70.3\\
  & +10.2& +13.2& +13.3\\\bottomrule
  \end{tabular}
%   }
  \caption{Comparison of Top-10 accuracy of BERT and ULR-BERT on different subsets of \textsc{GeoGranno}.}
  \label{rebuttal-2}
\end{table}

\subsection{Model Universality and Representation Consistency}
In previous evaluations on \textsc{GeoGranno}, our model has shown considerable improvement (10.2\% Top-10 Acc.) over BERT\textsubscript{BASE}. The task involves text matching between linguistic units at different levels where queries are sentences and labels are often phrases. Thus the performance on such task highly depends on the model's ability to uniformly deal with linguistic units of different granularities. 

In the following, we explore deeper details and interpretability of how our proposed objective act at different levels of linguistic units. Specifically, we group the dataset according to query length $|q|$ and the absolute difference between query length and Question length $abs(|q|-|Q|)$, respectively, to evaluate the universality of ULR-BERT and the consistency of the learned representations.

Results are shown in Table \ref{rebuttal-2}. As the length of the query increases, the performance of BERT drops sharply. Similarly, BERT is more sensible to the difference between query length and question length. In contrast, ULR-BERT is more stable when dealing with sequences of different lengths, which we speculate is due to the interaction between different levels of linguistic units in the pre-training procedure.

\section{Conclusion}
This work formally introduces universal language representation learning to enable unified vector operations among different language hierarchies. For such a purpose, we propose three highlighted ULR learning enhancement, including the newly designed training objective, Minimizing Symbol-vector Algorithmic Difference (MiSAD). In detailed model implementation, we extend BERT's pre-training objective to a more general level, which leverages information from sequences of different lengths in a comprehensive way. In addition, we provide a universal analogy dataset as a task-independent evaluation benchmark. Overall experimental results show that our proposed ULR model is generally effective in a broad range of NLP tasks including natural language question answering and so on.

\bibliographystyle{acl_natbib}
\bibliography{anthology,acl2021,ref}

\appendix
\section{Universal Analogy}
As a new task, universal representation has to be evaluated in a multiple-granular analogy dataset. In this section, we introduce the procedure of constructing different levels of analogy datasets based on Google's word analogy dataset. 

\subsection{Word-level analogy}
The goal of a word analogy task is to solve questions like ``$A$ is to $B$ as $C$ is to ?", which is to retrieve the last word from the vocabulary given the first three words. The objective can be formulated as maximizing the cosine similarity between the target word embedding and the linear combination of the given vectors:
\begin{align*}
     &d^* = \mathop{\arg\max}_{d^*} cosine(c+b-a, d)\\
     &cosine(u, v) = \frac{u\cdot v}{\|u\|\|v\|}
\end{align*}
where $a$, $b$, $c$, $d$ represent embeddings of the corresponding words and are all normalized to unit lengths.

We construct a closed-vocabulary analogy task based on Google's word analogy dataset through negative sampling. During evaluation, the model is expected to select the correct answer from 5 candidate words. 

\subsection{Phrase/Sentence-level analogy}
To investigate the arithmetic properties of vectors for higher levels of linguistic units, we present phrase and sentence analogy tasks based on the proposed word analogy dataset. Statistics are shown in Table \ref{statistics}.

\subsubsection{Semantic}
Semantic analogies can be divided into four subsets: ``capital-common", ``capital-world", ``city-state" and ``male-female". The first two sets can be merged into a larger dataset: ``capital-country", which contains pairs of countries and their capital cities; the third involves states and their cities; the last one contains pairs with gender relations. Considering GloVe's poor performance on word-level ``country-currency" questions ($<$32\%), we discard this subset in phrase and sentence-level analogies. Then we put words into contexts so that the resulting phrases and sentences also have linear relationships. For example, based on relationship
\begin{equation*}
    \textit{Athens} : \textit{Greece} :: \textit{Baghdad} : \textit{Iraq}, 
\end{equation*}
we select phrases and sentences that contain the word ``\textit{Athens}" from the English Wikipedia Corpus. We manually modify some words to ensure text coherence: ``\textit{He was hired as being professor of physics by the university of Athens.}" and create examples:\\[1.5pt]

\textit{hired by ... Athens} : \textit{hired by ... Greece} :: \textit{hired by ... Baghdad} : \textit{hired by ... Iraq}.\\[1.5pt]

\begin{table}[t]{}
    \small
  \centering
  \begin{tabular}{lrrrr}\toprule
    Dataset & \#p & \#q & \#c & \#l (p/s)\\\midrule
    capital-common	& 23	& 506	&5 &6.0/12.0\\
    capital-world	&116	&4524	&5 &6.0/12.0 \\
    city-state	&67&	2467&	5 &6.0/12.0\\
    male-female	&23	&506&	5 &4.1/10.1\\
    present-participle	&33	&1056&	2 &4.8/8.8\\
    positive-comparative	&37	&1322&	2 &3.4/6.1\\
    positive-negative &	29	&812	&2& 4.4/9.2\\\midrule
    All	&328	&11193	&- & 5.4/10.7\\\bottomrule
  \end{tabular}
  \caption{Statistics of our analogy datasets. \#p and \#q are the number of pairs and questions for each category. \#c is the number of candidates for each dataset. \#l (p/s) is the average sequence length in phrase/sentence-level analogy datasets.}
  \label{statistics}
\end{table}

However, we found that such a question is identical to word-level analogy for BOW methods like averaging GloVe vectors, because they treat embeddings independently despite the content and word order. To avoid lexical overlapping between sequences, we replace certain words and phrases with their synonyms and paraphrases, e.g.,\\[1.5pt]

\textit{hired by ... Athens} : \textit{employed by ... Greece} :: \textit{employed by ... Baghdad} : \textit{hired by ... Iraq}.\\[1.5pt]

\subsubsection{Syntactic}
We consider three typical syntactic analogies: Tense, Comparative and Negation, corresponding to three subsets: ``present-participle", ``positive-comparative", ``positive-negative", where the model needs to distinguish the correct answer from ``past tense", ``superlative" and ``positive", respectively. For example, given phrases\\[1.5pt]

\textit{Pigs are bright} : \textit{Pigs are brighter than goats} :: \textit{The train is slow}, \\[1.5pt]

\noindent
the model need to give higher similarity score to the sentence that contains ``\textit{slower}" than the one that contains ``\textit{slowest}". Similarly, we add synonyms and synonymous phrases for each question to evaluate the model ability of learning context-aware embeddings rather than interpreting each word in the question independently. For instance, ``\textit{pleasant}" $\approx$ ``\textit{not unpleasant}" and ``\textit{unpleasant}" $\approx$ ``\textit{not pleasant}". 

\end{document}